\documentclass{article}

\usepackage{microtype}
\usepackage{graphicx}
\usepackage{subcaption}
\usepackage{booktabs} 

\usepackage{hyperref}

\usepackage[preprint]{icml2026}

\usepackage{amsmath}
\usepackage{amssymb}
\usepackage{mathtools}
\usepackage{amsthm}
\usepackage{multirow}
\usepackage[table]{xcolor}
\usepackage{pifont}

\usepackage[capitalize,noabbrev]{cleveref}

\theoremstyle{plain}

\theoremstyle{definition}

\theoremstyle{remark}

\usepackage[textsize=tiny]{todonotes}

\begin{document}

\twocolumn[
  \icmltitle{From Physical Degradation Models to Task-Aware All-in-One Image Restoration}



  \icmlsetsymbol{equal}{*}

  \begin{icmlauthorlist}
    \icmlauthor{Hu Gao}{yyy}
    \icmlauthor{Xiaoning Lei}{comp}
    \icmlauthor{Xichen Xu}{yyy}
    \icmlauthor{Xingjian Wang}{sch}
    \icmlauthor{Lizhuang Ma}{yyy}
  \end{icmlauthorlist}

  \icmlaffiliation{yyy}{Shanghai Jiao Tong University, Shanghai, China}
  \icmlaffiliation{comp}{CATL, Ningde, China}
  \icmlaffiliation{sch}{Beijing Normal University, Beijing, China}

  \icmlcorrespondingauthor{Lizhuang Ma}{lzma@sjtu.edu.cn}

  \icmlkeywords{Machine Learning, ICML}

  \vskip 0.3in
]



\printAffiliationsAndNotice{}  

\begin{abstract}
All-in-one image restoration aims to adaptively handle multiple restoration tasks with a single trained model. Although existing methods achieve promising results by introducing prompt information or leveraging large models, the added learning modules increase system complexity and hinder real-time applicability. In this paper, we adopt a physical degradation modeling perspective and predict a task-aware inverse degradation operator for efficient all-in-one image restoration.
The framework consists of two stages. In the first stage, the predicted inverse operator produces an initial restored image together with an uncertainty perception map that highlights regions difficult to reconstruct, ensuring restoration reliability. In the second stage, the restoration is further refined under the guidance of this uncertainty map. The same inverse operator prediction network is used in both stages, with task-aware parameters introduced after operator prediction to adapt to different degradation tasks. Moreover, by accelerating the convolution of the inverse operator, the proposed method achieves efficient all-in-one image restoration.
The resulting tightly integrated architecture, termed OPIR, is extensively validated through experiments, demonstrating superior all-in-one restoration performance while remaining highly competitive on task-aligned restoration.
\end{abstract}

\section{Introduction}
Image restoration (IR) aims to recover a latent clean image $J$ from an observed degraded image $I$. The degradation process is generally formulated as:
\begin{equation}
I = \mathcal{D}_t(J)
\label{eq:general_degradation}
\end{equation}
where $\mathcal{D}_t(\cdot)$ denotes a task-dependent degradation operator and $t$ is the degradation type. The restoration objective is to estimate $J$ from $I$ under unknown and task-specific degradation mechanisms. IR is inherently ill-posed, as a single degraded image may correspond to multiple plausible solutions. Traditional methods~\cite{2011Single, 10558778} tackle this by introducing task-specific priors, modeling the degradation process, and applying inverse operations. While effective in some cases,  they depend heavily on strong assumptions about degradation factors. In real-world scenarios, where degradations are often uncertain or unknown, accurate modeling becomes challenging, leading to limited generalization and unstable performance.

Deep learning methods~\cite{PGH2Netisu2025prior,FSNet, VLUNetZeng_2025_CVPR} have significantly advanced IR by learning the statistical properties of natural images, enabling them to implicitly capture a wide range of priors and often outperform traditional approaches. According to how many degradation types a single model can accommodate, existing IR methods can be broadly divided into three categories: task-specific, task-aligned, and all-in-one. Task-specific approaches~\cite{PGH2Netisu2025prior, efderainguo2025efficientderain+} focus on individual restoration tasks by designing specialized models that are optimized for particular degradation characteristics. Task-aligned methods~\cite{FSNet, aclgu2025acl} pursue a more general framework by sequentially training a single network on datasets with different degradations, enabling it to support multiple restoration tasks. In contrast, all-in-one methods~\cite{potlapalli2023promptir,VLUNetZeng_2025_CVPR} train a unified model on mixed degradation types simultaneously, allowing one network to flexibly address a variety of restoration tasks.

\begin{figure}
    \centering
    \includegraphics[width=1\linewidth]{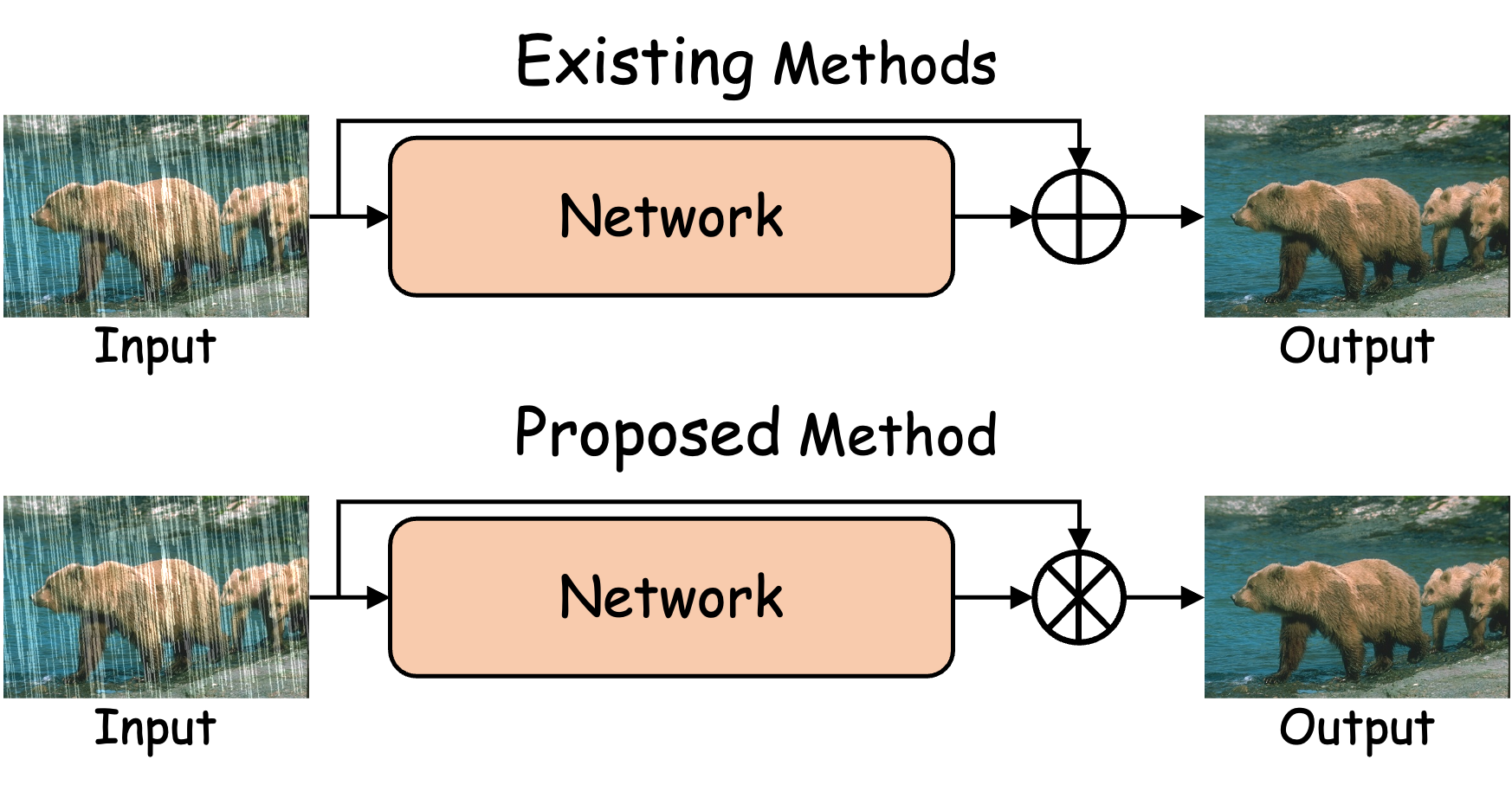}
    \caption{Mechanisms of existing methods vs. our method.}
    \label{fig:exam}
\end{figure}

Although existing all-in-one IR methods have demonstrated strong performance through the design of prompts or the adoption of large-scale models, the introduction of additional learning modules inevitably increases system complexity and limits their practicality in real-time applications. This observation motivates us to explore a more efficient alternative by learning task-aware inverse degradation operators directly from the degradation modeling process, aiming to achieve effective all-in-one IR while maintaining computational efficiency. As shown in Figure~\ref{fig:exam}, existing methods predict the residual image from the degraded input using deep learning models and obtain the final restored image by adding this residual to the degraded image. In contrast, our method predicts the inverse degradation operator and produces the restored image by convolving it with the degraded input. By accelerating this convolution process, efficient IR is achieved. 

\begin{figure}
    \centering
    \includegraphics[width=1\linewidth]{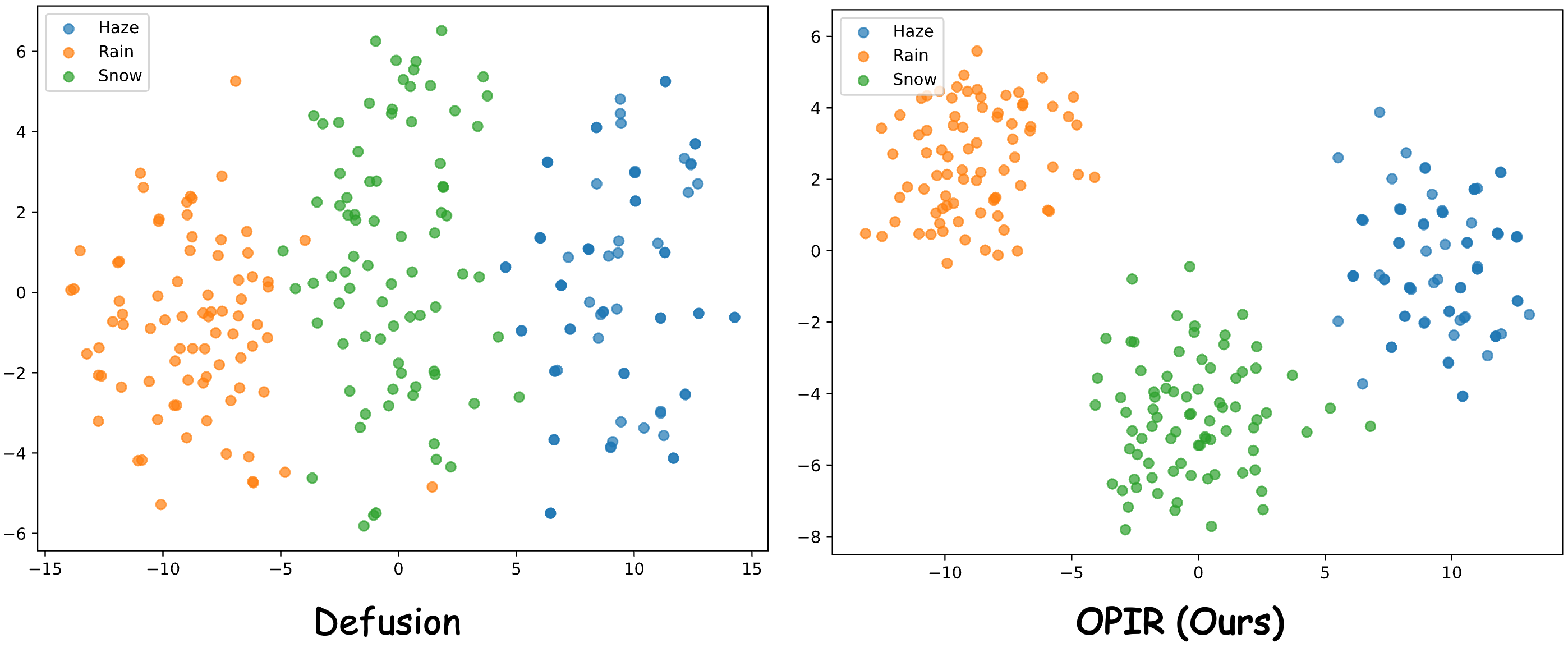}
    \caption{The figure shows t-SNE visualizations of degradation embeddings from OPIR (our method) and Defusion~\cite{DefusionLuo_2025_CVPR}, with our model displaying more distinct and well-separated clusters.}
    \label{fig:tsne}
\end{figure}

Specifically, our framework follows a two-stage design. In the first stage, the predicted inverse degradation operator produces an initial restored image along with an uncertainty perception map, which highlights regions that are difficult to reconstruct and ensures restoration reliability. In the second stage, this uncertainty map guides further refinement, enabling more precise recovery in challenging areas. The same inverse operator prediction network is used in both stages, with task-aware parameters introduced after operator prediction to adapt effectively to different degradation tasks. As shown in Figure~\ref{fig:tsne}, OPIR generates more distinct clusters, indicating a stronger ability to differentiate degradation types. Extensive experiments further demonstrate that our method achieves state-of-the-art performance in both all-in-one and task-aligned IR.

The main contributions of this work are:
\begin{enumerate}
	\item We propose OPIR for all-in-one image restoration, taking a physical degradation modeling approach and predicting a task-aware inverse degradation operator to achieve efficient restoration. Extensive experiments demonstrate its superior performance in both all-in-one and task-aligned IR.
    \item We design a two-stage framework where the predicted inverse degradation operator first generates an initial restored image with an uncertainty map to ensure reliability, and then this map guides further refinement for more accurate recovery in challenging regions.
	\item We design a task-aware inverse operator with adaptive parameters for different degradation tasks, which can be efficiently implemented via convolution accelerated
    
\end{enumerate}

\section{Related Works}

\subsection{Image Restoration}
Image restoration (IR) aims to recover high-quality images from degraded inputs, a challenging and inherently ill-posed problem. Traditional approaches~\cite{2011Single, 10558778} tackle this by introducing handcrafted priors to constrain the solution space. However, these priors rely heavily on expert knowledge and often exhibit limited adaptability, resulting in poor generalization across diverse degradation scenarios.

With the rapid development of deep learning in vision tasks, numerous data-driven IR methods~\cite{LSSRgao2024learning,adarevD10656920,FSNet,aclgu2025acl,VLUNetZeng_2025_CVPR} have been proposed, leveraging statistical regularities of natural images to implicitly learn priors and surpass traditional methods. Depending on how many degradation types a single model can handle, these methods are broadly classified into task-specific, task-aligned, and all-in-one approaches.

Task-specific methods~\cite{PGH2Netisu2025prior, efderainguo2025efficientderain+} design dedicated models for individual degradations, often customized to the target dataset.  ALGNet~\cite{ALGgao2024learning} introduces local feature extraction to reduce pixel forgetting for precise deblurring, XYScanNet~\cite{liu2024xyscannet} uses alternating intra- and inter-slice scanning to capture spatial dependencies, and EfDeRain+~\cite{efderainguo2025efficientderain+} formulates deraining as predictive filtering, avoiding complex rain modeling. Diffusion-based approaches, such as UPID-EDM~\cite{upid10.1145/3664647.3680560}, leverage vision-language priors to remove degradations while preserving structure, while PGH$^2$Net~\cite{PGH2Netisu2025prior} combines bright/dark channel priors with histogram equalization for hierarchical dehazing. Diff-Unmix~\cite{diffunmix10656884} achieves self-supervised denoising through spectral decomposition and conditional diffusion.

Task-aligned methods~\cite{FSNet, aclgu2025acl} aim to build general networks capable of handling multiple degradations by sequential training across diverse datasets. MambaIR~\cite{guo2025mambair} and MambaIRV2~\cite{guo2025mambairv2} introduce multi-directional unfolding and non-causal modeling to enhance spatial and state-space representations. SFNet~\cite{SFNet} and FSNet~\cite{FSNet} employ dynamic frequency selection to focus on the most informative components, while ACL~\cite{aclgu2025acl} exploits the equivalence between linear attention and state-space models to unlock linear attention for restoration. MHNet~\cite{gao2025mixed} adopts a mixed hierarchical design to generate restorations with richer textures and finer structural details.

All-in-one methods~\cite{potlapalli2023promptir,VLUNetZeng_2025_CVPR} extend this concept further, training a single network to handle multiple degradation types simultaneously. PromptIR~\cite{potlapalli2023promptir} uses lightweight prompt modules for direct restoration, CAPTNet~\cite{CAPTNet10526271} leverages data-component changes through prompt learning, and NDR~\cite{NDR10680296} captures shared degradation characteristics via neural representations. Large models further enhance all-in-one IR. VLU-Net~\cite{VLUNetZeng_2025_CVPR} uses vision-language features to automatically identify degradation-aware keys, AutoDIR~\cite{autodir10.1007/978-3-031-73661-2_19} combines VLM guidance with latent diffusion for structure-consistent restoration, AdaIR~\cite{cui2025adair} separates degradation from content across spatial and frequency domains, and Defusion~\cite{DefusionLuo_2025_CVPR} applies visual instruction-guided degradation diffusion. Perceive-IR~\cite{Perceive-IR10990319} introduces a two-stage design that identifies both degradation types and fine-grained severity levels, achieving strong transferability across restoration tasks.
Although existing all-in-one IR methods achieve strong performance using prompts or large-scale models, added learning modules increase system complexity and limit real-time applicability. In this paper, we take a physical degradation modeling approach and predict a task-aware inverse degradation operator for efficient all-in-one image restoration.

\subsection{Kernel-Predicted Restoration}
Kernel-predicted techniques have been widely used in image restoration due to the local coherence, continuity, and structural properties of natural images. Traditional methods rely on handcrafted kernels and priors to process images locally, including adaptive~\cite{tada1580482}, recursive~\cite{trec655423}, bilateral~\cite{tmulb4664614}, and collaborative 3-D transform-domain approaches~\cite{t3d4271520}. While effective in certain cases, these methods often struggle with complex degradations such as rain streaks, where bilateral kernels may blur the image and guided kernels can introduce artifacts. With the rise of deep learning, predictive kernel methods have emerged as a powerful alternative. Kernel Prediction Networks (KPN)~\cite{kpnmildenhall2018burst} learn per-pixel, image-adaptive kernels through deep convolutional networks, enabling flexible, data-driven restoration. These approaches have been successfully applied to various image restoration tasks~\cite{kpn210637923,kpn3guo2021jpgnet,kpn49578628}. However, most existing methods overlook multi-scale modeling, and the effectiveness of the predicted kernels is seldom explicitly evaluated, which can result in distorted details. To alleviate this issue, EfDeRain+~\cite{efderainguo2025efficientderain+} introduces a weight-sharing multi-scale dilated filtering strategy, but it remains task-specific and cannot accommodate multiple restoration tasks within a unified framework. To overcome these limitations, we propose a task-aware kernel prediction framework that incorporates multi-scale uncertainty modeling and task-aware parameters, enabling efficient all-in-one image restoration.

\section{Method}

\subsection{Problem Definition}
As shown in Eq.~\ref{eq:general_degradation}, the restoration objective is to estimate $J$ from $I$ under unknown and task-specific degradation mechanisms. The modeling of degradation varies across different restoration tasks.
Deraining is typically modeled as an additive process:
\begin{equation}
I = J + \sum_{k=1}^K R_k
\end{equation}
where $R_k$ denotes individual rain streak components.
Desnowing explicitly accounts for occlusion effects through a masking mechanism:
\begin{equation}
I = M \odot S + (1 - M) \odot J
\end{equation}
where $M$ represents the snow mask and $S$ denotes snow particles.
Dehazing follows the atmospheric scattering model:
\begin{equation}
I(x) = J(x),t(x) + A\big(1 - t(x)\big)
\end{equation}
where $t(x)$ is the transmission map and $A$ denotes the global atmospheric light.
Despite their different formulations, these degradation models can be unified as:
\begin{equation}
I = H_t J + b_t
\end{equation}
where $H_t$ represents a task-dependent degradation operator and $b_t$ denotes a bias term.
Under this unified view, deraining, desnowing, and dehazing can all be interpreted as inverse problems with unknown degradation operators.

Although the degradation operators differ across tasks, their corresponding bias terms are often strongly correlated. Therefore, we adopt a joint prediction strategy.
Instead of explicitly estimating $H_t$ and $b_t$, we directly learn a task-aware inverse mapping:
\begin{equation}
\hat{J} = \mathcal{G}(I; \theta_t)
\label{eq:prek}
\end{equation}
where $\mathcal{G}(\cdot)$ denotes a shared restoration backbone modulated by task-aware parameters $\theta_t$.
From an operator-learning perspective, the model learns a family of inverse operators $\mathcal{G}_t \approx \mathcal{D}_t^{-1}$.
Accordingly, image restoration can be formulated as the following optimization problem:
\begin{equation}
\min\left|| \mathcal{G}(I; \theta_t) - J \right||
\label{eq:joint_optimization}
\end{equation}

\begin{figure}
    \centering
    \includegraphics[width=1\linewidth]{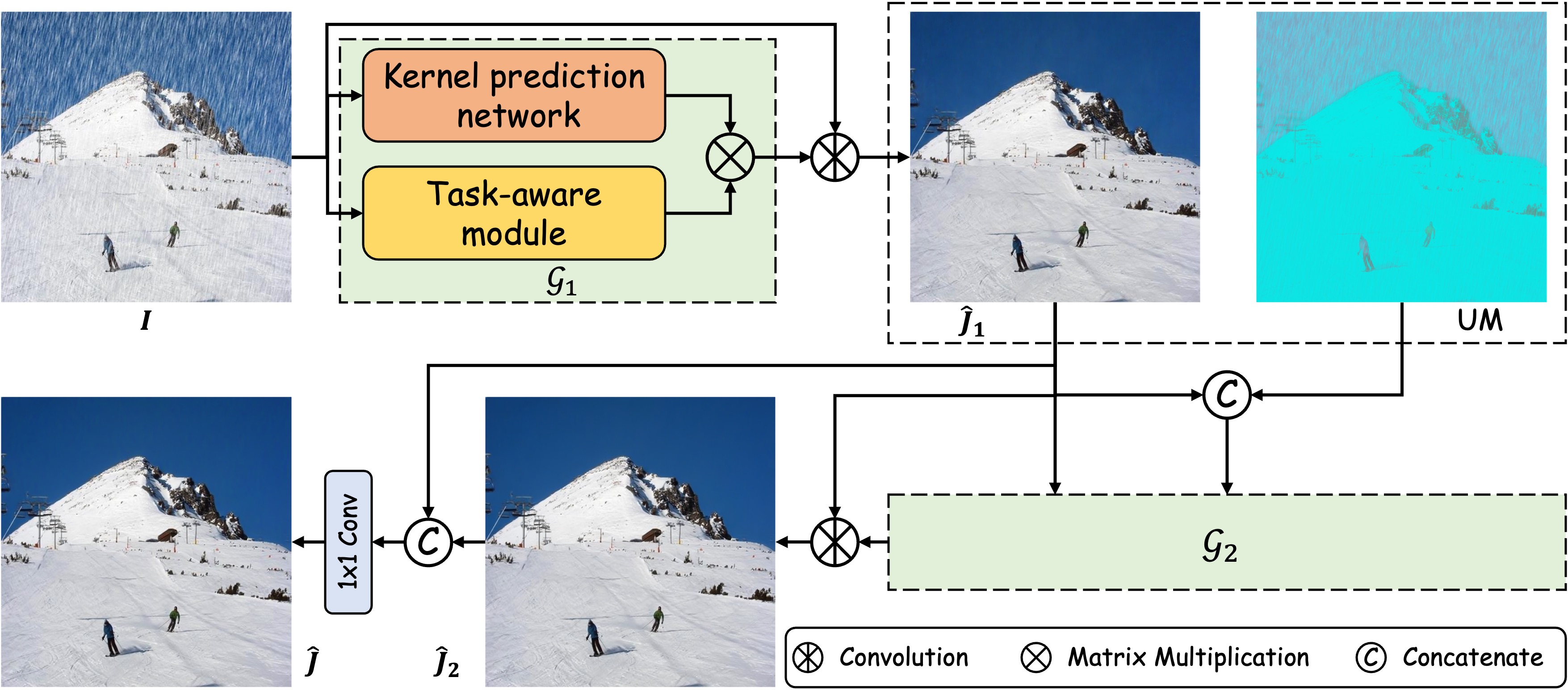}
    \caption{The overall architecture of the proposed OPIR is composed of two stages.}
    \label{fig:network}
\end{figure}

To further enhance restoration performance, as shown in Figure~\ref{fig:network}, we design a two-stage strategy.
In the first stage, the model predicts not only the inverse mapping but also an uncertainty estimation map.
In the second stage, this uncertainty map is leveraged to focus on severely degraded regions, enabling an all-in-one restoration framework across diverse degradation types.
Accordingly, Eq.~\eqref{eq:prek} can be rewritten as:
\begin{equation}
\hat{J} = \mathcal{G}_2\big(\mathcal{G}_1(I; \theta_{t1}); \theta_{t2}\big)
\end{equation}

In this paper, our objective is to learn $\mathcal{G}_1(\cdot)$ and $\mathcal{G}_2(\cdot)$. The functions $\mathcal{G}_1(\cdot)$ and $\mathcal{G}_2(\cdot)$ share the same network architecture and are defined as:
\begin{equation}
\begin{aligned}
\mathcal{G}_1(I; \theta_t) &= \mathrm{TAIOM}(I) 
\\
&= \big( \mathrm{KPN}(I) \otimes \mathrm{TAM}(I) \big) \textcircled{\textasteriskcentered} I
\label{eq:gnet}
\end{aligned}
\end{equation}
where TAIOM denotes the task-aware inverse operator modeling. Here, KPN refers to the kernel prediction network implemented with a U-Net~\cite{chen2022simple} architecture, and TAM represents the task-aware module.

\subsection{Task-Aware Inverse Operator Modeling}
Different restoration tasks correspond to distinct physical degradation processes, requiring specialized inverse operators. Rather than learning independent operators for each task, we introduce a task-aware modulation mechanism to adapt a shared predicted operator. Specifically, a lightweight task-aware module $\mathcal{T}(\cdot)$ modulates the base operator using task-specific parameters $\theta_t$:

\begin{equation}
\tilde{K} = \mathcal{T}(K; \theta_t) = KPN(I) \otimes TAM(I)
\end{equation}

where $KPN(\cdot)$ denotes the kernel prediction network and $TAM(\cdot)$ the task-aware module. This design allows a single backbone to predict operators while maintaining task-specific adaptability with minimal overhead.

To improve restoration reliability, we estimate per-pixel reconstruction uncertainty from the predicted operator. For pixel $p$, the uncertainty score is computed as the mean absolute value of kernel weights in its local neighborhood $\mathcal{N}(p)$:

\begin{equation}
UM(p) = \frac{1}{k^2} \sum_{q \in \mathcal{N}(p)} \left| K_p(q) \right|
\end{equation}

yielding an uncertainty map $UM \in \mathbb{R}^{H \times W}$ that highlights regions difficult to reconstruct and serves as guidance in the refinement stage.
Based on this formulation, we adopt a two-stage inverse operator modeling strategy. In the first stage, the task-aware operator produces an initial restoration $\hat{J}_1$ along with the uncertainty map:

\begin{equation}
\hat{J}_1 = \tilde{K}_1 \textcircled{\textasteriskcentered} I
\end{equation}

In the second stage, the same network refines the restoration conditioned on the intermediate result and uncertainty map:

\begin{equation}
\begin{aligned}
\hat{J}_2 &= \tilde{K}_2 \textcircled{\textasteriskcentered} \hat{J}_1, \\
\tilde{K}_2 &= \mathcal{T}(K_2; \theta_{t2}) = KPN(\hat{J}_1 \textcircled{c} UM) \otimes TAM(\hat{J}_1)
\end{aligned}
\end{equation}

By explicitly modeling task-awareness and uncertainty-guided refinement, the proposed TAIOM captures degradation-specific characteristics while generalizing across diverse tasks.

\begin{table*}
\centering
\caption{Quantitative results in the all-in-one setting with task-specific, task-aligned, and all-in-one  IR methods. }
\label{tb:all}
    \resizebox{\linewidth}{!}{
\begin{tabular}{c|c|cccccc|cc}
    \hline
   \multirow{2}{*}{Type} & \multirow{2}{*}{Methods} & \multicolumn{2}{c}{Deraining} & \multicolumn{2}{c}{Desnowing} & \multicolumn{2}{c}{Dehazing}  & \multicolumn{2}{|c}{Average} 
    \\
   & &PSNR $\uparrow$ &  SSIM $\uparrow$  &PSNR $\uparrow$ &SSIM $\uparrow$ & PSNR $\uparrow$&SSIM $\uparrow$ &PSNR $\uparrow$ & SSIM $\uparrow$
    \\
    \hline\hline
   \multirow{4}{*}{Specific} 
   &MSP-Former~\cite{mspformer10095605} & 29.25 & 0.906 & 32.81 & \underline{0.955} & 32.89 & 0.970 &  31.65 & 0.944
     \\
   &EfDeRain+~\cite{efderainguo2025efficientderain+} & 31.18 & \underline{0.912} &  29.96 & 0.938 & 33.25 & 0.971 & 31.46 & 0.940
     \\
     &PGH$^2$Net~\cite{PGH2Netisu2025prior}& 29.89 & 0.907 & 32.18 & 0.943 & 35.55 & 0.983 & 32.21 & 0.944
     \\
     &ALGNet~\cite{ALGgao2024learning}& 29.32 &0.905 & 31.56 & 0.944 & 33.11 & 0.977 & 31.33 & 0.942
     \\
     \hline
    \multirow{4}{*}{Aligned}
    &FSNet~\citep{FSNet} & 30.85 & 0.909 & 32.88 & 0.947 & 34.82 & 0.981 & 32.85 & 0.946
    \\
    &MHNet~\citep{gao2025mixed} & 30.69 & 0.906 & 32.03 & 0.943 & 34.59 & 0.980 & 32.44 & 0.943
     \\
     &PPTformer~\cite{pptformerwang2025intra} & 30.88 & 0.910 & 32.80 & 0.945 & 34.66 & 0.980 & 32.78 & 0.945
     \\
    &ACL~\cite{aclgu2025acl} &31.11 & \underline{0.912} & 32.91 & 0.948 & 34.81 & 0.981 & 32.94 & 0.947
     \\
     \hline
      \multirow{5}{*}{All-in-One}
      &AdaIR~\cite{cui2025adair}& \underline{31.22} & 0.911 & 32.99 & 0.949 & 35.31 & 0.982 & 33.17 & 0.947
      \\
      &VLU-Net~\cite{VLUNetZeng_2025_CVPR}& 31.07 & 0.910 & 33.02 & 0.952 & 35.43 & \underline{0.984} & 33.17 & \underline{0.949}
      \\
      &Perceive-IR~\cite{Perceive-IR10990319}& 31.13 & 0.911 & 33.11 & 0.952 & 35.35 & 0.983 & 33.20 & \underline{0.949}
        \\
        &Defusion~\cite{DefusionLuo_2025_CVPR}& 30.98 & 0.910 & \underline{33.29} & 0.953 & \underline{35.52} & \textbf{0.985}& \underline{33.26} & \underline{0.949}
        \\
      &\cellcolor{gray!20}\textbf{OPIR(Ours)} &\cellcolor{gray!20}\textbf{31.96} 
      &\cellcolor{gray!20}\textbf{0.914} &\cellcolor{gray!20}\textbf{34.14}  &\cellcolor{gray!20}\textbf{0.959} 
      &\cellcolor{gray!20}\textbf{35.96}
      &\cellcolor{gray!20}\underline{0.984} &\cellcolor{gray!20}\textbf{34.02} 
      &\cellcolor{gray!20}\textbf{0.952}
    \\
    \hline
\end{tabular}}
\end{table*}

Furthermore, for degradations that span large regions,  a small receptive field may fail to capture sufficient clean context, whereas larger kernels can leverage distant pixels to enhance reconstruction quality. This observation naturally motivates a multi-scale filtering formulation. Following~\cite{efderainguo2025efficientderain+}, we employ a  multi-scale filtering strategy that decouples kernel size from kernel parameters, significantly reducing computational complexity. To recover a clean image $\hat{J}$ from an observed degraded image $I \in \mathbb{R}^{H \times W \times C}$, instead of predicting separate kernels $K^{(s)}$ for each scale $s$, we define a single shared base kernel and reuse it across multiple scales via dilation:
\begin{equation}
K \in \mathbb{R}^{H \times W \times 3 \times 3}
\end{equation}

\begin{equation}
K^{(s)}_p(\Delta) = K_p(\delta), \quad \Delta = s \cdot \delta, \quad \delta \in \{-1,0,1\}^2
\end{equation}
where positions not covered by this mapping are set to zero. The output at scale $s$ for pixel $p$ is:

\begin{equation}
\hat{J}^{(s)}(p) = \sum_{\delta \in \{-1,0,1\}^2} K_p(\delta) \cdot I(p + s \delta)
\end{equation}

To further reduce computation, we note that kernel weights are shared across scales. Let $\Delta_s = \{s \cdot \delta \mid \delta \in \{-1,0,1\}^2\}$ denote sampled offsets, and $I_\text{sampled} \in \mathbb{R}^{H \times W \times 9 \times S}$ the tensor of sampled image values for $S$ scales. Then the multi-scale convolution can be written as:

\begin{equation}
\hat{J}_p^{(s)} = \sum_{i=1}^{9} K_p(\delta_i) \cdot I_\text{sampled}(p, i, s)
\end{equation}

reducing per-pixel complexity from $\mathcal{O}(HW(2s+1)^2)$ to $\mathcal{O}(9HW)$, independent of $S$, and avoiding explicit materialization of $K^{(s)}$.
Finally, we adopt a learnable fusion  $\alpha^{(s)}_p$ across scales:
\begin{equation}
\hat{I}_p = \sum_{s=1}^{S} \alpha_p^{(s)} \cdot \hat{I}^{(s)}_p, \quad \sum_{s=1}^S \alpha^{(s)}_p = 1, \quad \alpha^{(s)}_p \ge 0
\end{equation}

This approach enables adaptive selection of the effective receptive field per pixel while maintaining low computational footprint. Compared to~\cite{efderainguo2025efficientderain+}, our method generates all scale offsets in a single loop, performs multi-scale convolution in parallel, and compresses redundant computation with learnable fusion weights.

\subsection{Task-Aware Module}

Different image restoration tasks correspond to distinct degradation processes, requiring task-specific adaptation of the predicted inverse operator. To achieve this efficiently, we introduce a task-aware module (TAM) that modulates a shared base kernel $K$ using lightweight, task-specific parameters $\theta_t$. 
The TAM also facilitates spatially-varying modulation to emphasize different regions depending on the task. Let $p$ index a pixel and $\mathcal{N}(p)$ its local neighborhood. Then the modulated kernel for pixel $p$ is:

\begin{equation}
\tilde{K}_p(q) = K_p(q) \cdot m_p(q; \theta_t), \quad q \in \mathcal{N}(p),
\end{equation}

where $m_p(q; \theta_t)$ is the task-aware modulation weight predicted by TAM. The restored pixel is then obtained via local convolution:

\begin{equation}
\hat{J}_p = \sum_{q \in \mathcal{N}(p)} \tilde{K}_p(q) \cdot I(q).
\end{equation}

To jointly handle multiple tasks, we introduce a vector of task embeddings $\mathbf{e}_t \in \mathbb{R}^d$, and define the TAM output as a function of both the local kernel and the task embedding:

\begin{equation}
m_p(q; \theta_t) = g_\text{TAM}(K_p(q), \mathbf{e}_t),
\end{equation}
where $g_\text{TAM}(\cdot)$ is implemented as a lightweight MLP. This design allows the operator to adapt dynamically to the degradation type while maintaining a shared backbone, reducing memory and computation overhead compared to learning independent kernels for each task.

\begin{figure*} 
    \centerline{\includegraphics[width=1\linewidth]{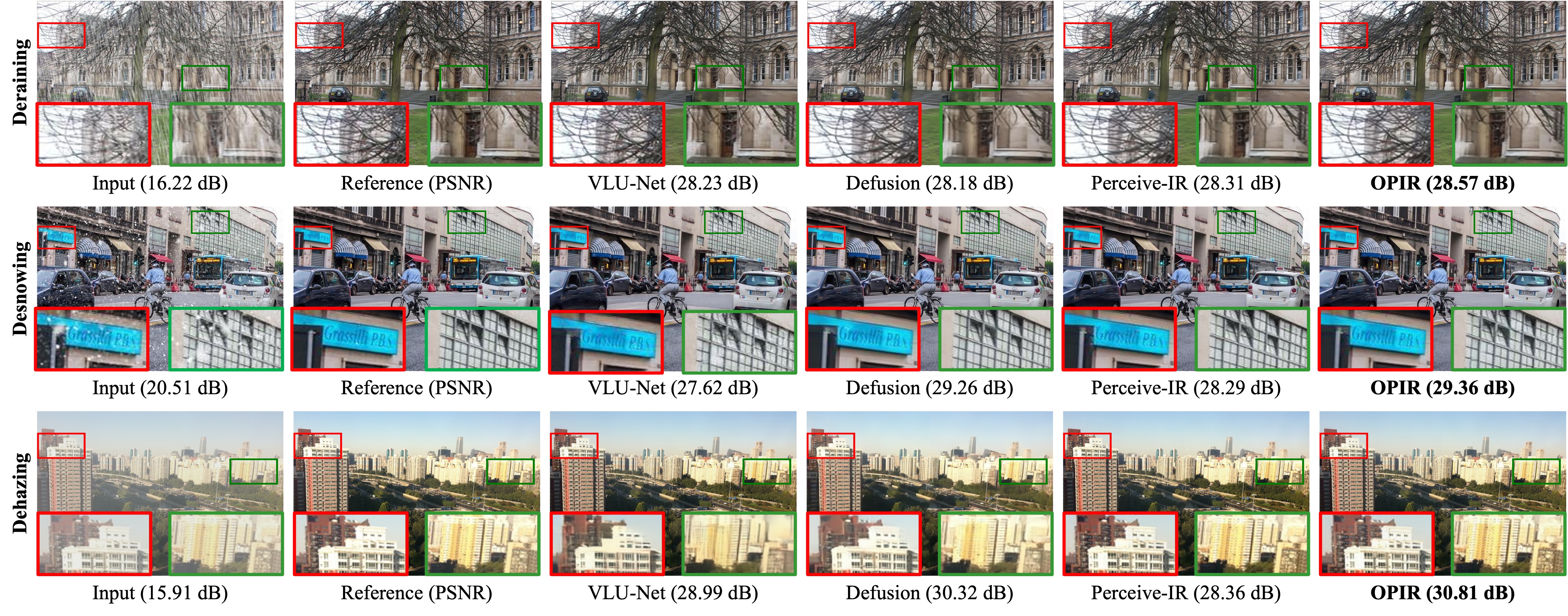}}
	\caption{Qualitative results under the all-in-one experimental setup. Our OPIR recovers finer details in the reconstructed images.}
 \label{fig:all-in-one}
\end{figure*}

\section{Experiments}
\label{sec:exp}
In this section, we first describe the experimental setup, followed by qualitative and quantitative comparison results. Finally, we present ablation studies to validate the effectiveness of our approach. \textbf{Due to page limits, more experiments we show in the appendix~\ref{sec:me}.}

\subsection{Experimental Setup}
We conducted experiments under both the all-in-one and task-aligned settings. \textbf{Owing to page limitations, detailed information about the datasets and training procedures is provided in Appendix~\ref{sec:mes}.}

\textbf{Datasets.} For the task-aligned setting, we train the deraining model using 13,712 paired clean–rain images collected from multiple datasets~\cite{Rain100,Test100,8099669,7780668}. The trained OPIR is evaluated on several benchmark test sets, including Rain100H~\cite{Rain100}, Rain100L~\cite{Rain100}, Test100~\cite{Test100}, and Test1200~\cite{DIDMDN}. For desnowing evaluation, we adopt the Snow100K~\cite{desnownet}, SRRS~\cite{JSTASRchen2020jstasr}, and CSD~\cite{HDCW-Netchen2021all} datasets. 
For dehazing, we assess the performance of our OPIR on daytime synthetic datasets from RESIDE~\cite{RESIDEli2018benchmarking}, which include the indoor training set (ITS), the outdoor training set (OTS), and the synthetic objective testing set (SOTS). 
For the all-in-one setting, we combine the above image restoration datasets to construct a  training set. During evaluation, we use Test100~\cite{Test100} for deraining, Snow100K~\cite{desnownet} for desnowing, and SOTS-Outdoor~\cite{RESIDEli2018benchmarking} for dehazing.

\subsection{Experimental Results}

\subsubsection{All-in-one Setting}

Table~\ref{tb:all} provides a comprehensive comparison among task-specific, task-aligned, and all-in-one image restoration methods under the all-in-one evaluation setting. The results clearly demonstrate the effectiveness and practical value of the proposed OPIR in terms of performance gains and robustness across diverse restoration tasks.

Compared with task-specific methods, OPIR exhibits significantly better generalization. Although task-specific approaches such as PGH$^2$Net~\cite{PGH2Netisu2025prior} and EfDeRain+~\cite{efderainguo2025efficientderain+} achieve strong performance on their respective target degradations, their performance degrades noticeably on other tasks, resulting in limited average scores. In contrast, OPIR surpasses the best task-specific method by +1.81 dB in average PSNR demonstrating.

Compared with task-aligned methods, which are trained sequentially across different datasets, OPIR consistently achieves higher accuracy. For example, OPIR improves upon the strongest task-aligned baseline ACL~\cite{aclgu2025acl} by +1.08 dB in average PSNR. This indicates that explicit task-aware operator modeling is more effective than sequential alignment in capturing the intrinsic differences among degradation types.

Compared with state-of-the-art all-in-one methods, OPIR still delivers clear and consistent improvements. Specifically, OPIR outperforms Defusion~\cite{DefusionLuo_2025_CVPR}, the strongest competing all-in-one baseline, by +0.76 dB in average PSNR. Moreover, OPIR achieves the best PSNR on all three tasks, +0.74 dB over the best baseline on deraining, +0.85 dB on desnowing, and +0.44 dB on dehazing. These consistent gains across tasks highlight the advantage of modeling task-aware inverse degradation operators over prompt-based or diffusion-based designs.
As shown in Figure~\ref{fig:all-in-one}, our model produces restored images that are sharper and visually closer to the ground truth than  others.

\subsubsection{Task-aligned Setting}
To demonstrate that the proposed OPIR is effective not only for all-in-one restoration but also for task-specific scenarios, we conduct experiments on three representative image restoration tasks: image deraining, image denoising, and image dehazing.

\begin{table*}
\centering
\caption{Image deraining results in the task-aligned setting.}
\label{tb:derain}
    \resizebox{\linewidth}{!}{
\begin{tabular}{c|cccccccc||cc}
    \hline
    \multicolumn{1}{c|}{} & \multicolumn{2}{c}{Test100}  & \multicolumn{2}{c}{Test1200} & \multicolumn{2}{c}{Rain100H} & \multicolumn{2}{c||}{Rain100L} & \multicolumn{2}{c}{Average} 
    \\
   Methods &PSNR $\uparrow$ &  SSIM $\uparrow$  & PSNR $\uparrow$ & SSIM $\uparrow$ &PSNR $\uparrow$ &SSIM $\uparrow$ & PSNR $\uparrow$&SSIM $\uparrow$ &PSNR $\uparrow$ & SSIM $\uparrow$
    \\
    \hline\hline
    FSNet~\citep{FSNet} &31.05&0.919 &33.08&0.916 & 31.77&0.906 &38.00 & 0.972 &33.48 &0.928
    \\
    MHNet~\citep{gao2025mixed} &31.25 &0.901 &\underline{33.45} &0.925 &31.08 &0.899 &\underline{40.04} &\underline{0.985} &33.96 &0.928
     \\
     PPTformer~\cite{pptformerwang2025intra} & 31.48 & \underline{0.922} & 33.39 & 0.911 &  31.77 & 0.907
     & 39.33 &0.983 & 33.99 & 0.931
     \\
 ACL~\cite{aclgu2025acl} &\underline{31.51} & 0.914 & 33.27 &\underline{0.928}  & 32.22 & \underline{0.920} & 39.18 &0.983  & 34.05 & \underline{0.936}
     \\
     EfDeRain+~\cite{efderainguo2025efficientderain+} &31.10 &0.911 &33.12 & 0.925 & \textbf{34.57} &\textbf{0.957} & 39.03 & 0.972 &\underline{34.46} & \textbf{0.941}
     \\
      \hline
    \rowcolor{gray!20}  \textbf{OPIR(Ours)}  & \textbf{33.15}	&\textbf{0.923}	&\textbf{35.19}	&\textbf{0.941}	&\underline{32.38}	&0.913	&\textbf{41.01}	&\textbf{0.986}	&\textbf{35.43}	&\textbf{0.941}
    \\
    \hline
\end{tabular}}
\end{table*}

\begin{figure*} 
    \centerline{\includegraphics[width=1\linewidth]{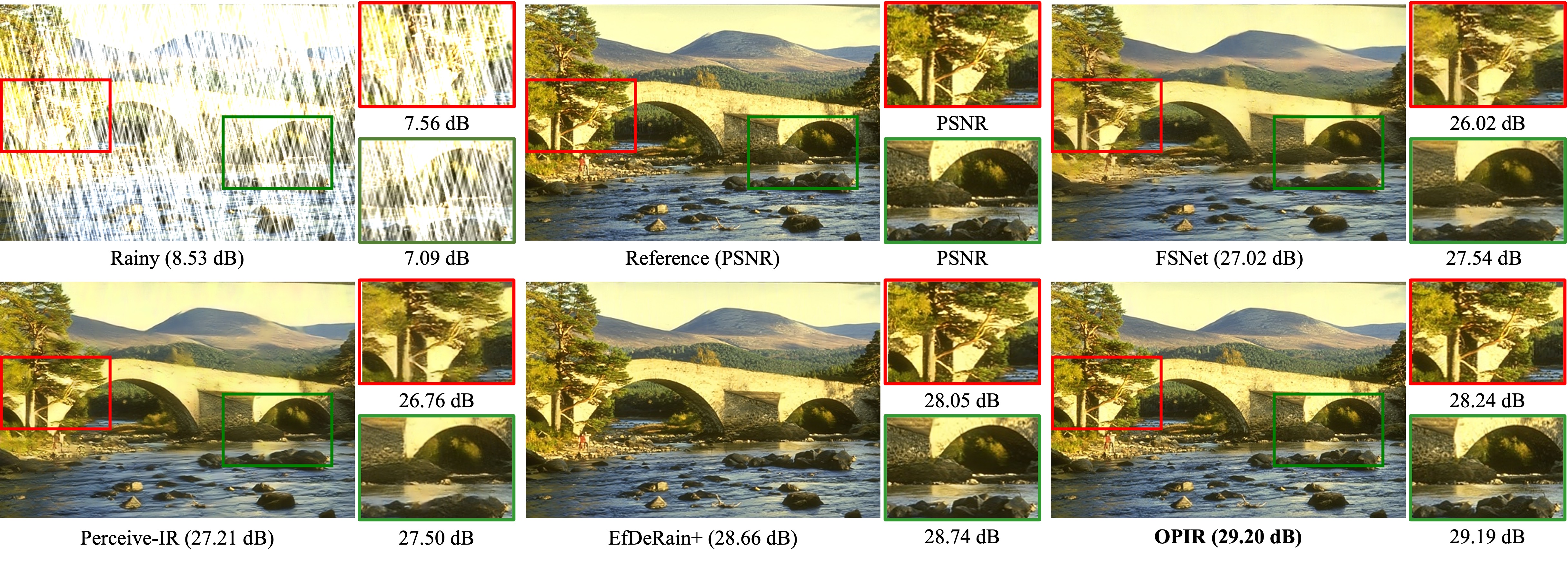}}
	\caption{Qualitative results under the task-aligned experimental setup. Our OPIR is able to reconstruct finer and sharper details.}
 \label{fig:srain}
\end{figure*}

\textbf{Image Deraining.} Following the protocol in prior work~\cite{gao2025mixed}, we evaluate PSNR and SSIM metrics on the Y channel of the YCbCr color space for the image deraining task. Table~\ref{tb:derain} reports the quantitative results of image deraining under the task-aligned setting. Our proposed OPIR consistently outperforms all compared methods across most datasets. Specifically, on Test100~\cite{Test100} and Test1200~\cite{MSPFN}, OPIR achieves 33.15 dB and 35.19 dB in PSNR, respectively, surpassing previous state-of-the-art methods by a clear margin. On Rain100L~\cite{Rain100}, OPIR reaches 41.01 dB PSNR, demonstrating superior capability in handling light rain conditions. Although EfDeRain+~\cite{efderainguo2025efficientderain+} attains the highest scores on Rain100H~\cite{Rain100}, our method still maintains competitive performance while achieving the best average PSNR (35.43 dB) and SSIM (0.941) across all datasets. These results indicate that OPIR not only effectively restores diverse rain streak patterns but also provides stable and robust performance in various deraining scenarios.
Figure~\ref{fig:srain} illustrates that the images restored by our OPIR effectively reduce color distortion compared to other state-of-the-art methods. Moreover, our approach is able to reconstruct finer and sharper details in the degraded regions.

\begin{table}
    \centering
        \caption{Image  desnowing results in the task-aligned setting.}
    \label{tab:snow}
    \resizebox{\linewidth}{!}{
    \begin{tabular}{c|cccccc}
    \hline
    \multicolumn{1}{c|}{} & \multicolumn{2}{c}{CSD}  & \multicolumn{2}{c}{SRRS} & \multicolumn{2}{c}{Snow100K}
    \\
   Methods & PSNR $\uparrow$ & SSIM $\uparrow$  & PSNR $\uparrow$ & SSIM $\uparrow$ & PSNR $\uparrow$ & SSIM $\uparrow$
   \\
   \hline
   \hline
 NAFNet~\cite{chen2022simple} &33.13 &0.96 &29.72 &0.94 &32.41 &\underline{0.95}
\\
FocalNet~\cite{focalnetcui2023focal} &37.18 &\textbf{0.99} &31.34 &\textbf{0.98} &33.53 &\underline{0.95}
\\
MSP-Former~\cite{mspformer10095605} & 33.75 &0.96 &30.76 &\underline{0.95} &33.43& \textbf{0.96}
\\
IRNeXt~\cite{IRNeXt} &\underline{37.29} &\textbf{0.99} &\underline{31.91} &\textbf{0.98} &33.61 &\underline{0.95}
\\
PEUNet~\cite{PEUNet10830558}&37.28 &0.97 &31.89 &\textbf{0.98} &\underline{34.11} &\textbf{0.96}
\\
\hline
\rowcolor{gray!20} \textbf{OPIR(Ours)} &\textbf{37.74}	&\underline{0.98}  &\textbf{33.10}	&\textbf{0.98}		&\textbf{34.58}	&\textbf{0.96}

         \\
         \hline
    \end{tabular}}
\end{table}

\textbf{Image Desnowing.}
Table~\ref{tab:snow} presents quantitative comparisons of image desnowing performance on three benchmark datasets: CSD, SRRS, and Snow100K. Our proposed OPIR consistently achieves the highest PSNR across all datasets, with 37.74 dB on CSD, 33.10 dB on SRRS, and 34.58 dB on Snow100K. In particular, OPIR achieves  a 0.45 dB improvement on the CSD dataset~\cite{HDCW-Netchen2021all}, an impressive 2.07 dB gain on the SRRS dataset~\cite{JSTASRchen2020jstasr} and a 0.97 dB improvement on the  Snow100K dataset~\cite{desnownet} when compared with IRNeXt~\cite{IRNeXt}. These results highlight the model’s superior capability in handling diverse snow patterns and background complexities. Furthermore, relative to MSP-Former~\cite{mspformer10095605}, a task-specific architecture tailored for image desnowing, OPIR exhibits substantial quantitative gains of 1.15 dB, 2.34 dB, and 3.99 dB on the Snow100K, SRRS, and CSD datasets, respectively. This demonstrates that our adaptive multi-degradation design not only generalizes effectively to snow removal tasks but also achieves state-of-the-art restoration quality across different data distributions. These results demonstrate that OPIR not only excels in desnowing across multiple datasets but also provides robust and reliable performance.

\begin{table}
    \centering
        \caption{Image dehazing results in the task-aligned setting.}
    \label{tab:sot}
    \resizebox{\linewidth}{!}{
    \begin{tabular}{c|cccc}
    \hline
    \multicolumn{1}{c|}{} & \multicolumn{2}{c}{SOTS-Indoor}  & \multicolumn{2}{c}{SOTS-Outdoor} 
    \\
   Methods & PSNR $\uparrow$ & SSIM $\uparrow$  & PSNR $\uparrow$ & SSIM $\uparrow$ 
   \\
   \hline
   \hline
SFNet~\cite{SFNet} & 41.24 &\underline{0.996} &\underline{40.05} &\textbf{0.996} 
\\
IRXNext~\cite{IRNeXt}&41.21 &\underline{0.996} &39.18 &\textbf{0.996}       
\\
DEA-Net-CR~\cite{deanetchen2024dea} & 41.31&0.995 & 36.59 & 0.990 
\\
 Defusion~\cite{DefusionLuo_2025_CVPR} & 41.65&0.995 & 37.41 & \underline{0.993}
\\
PGH$^2$Net~\cite{PGH2Netisu2025prior}&\underline{41.70} &\underline{0.996}&37.52 &0.989
         \\
         \hline
       \rowcolor{gray!20} \textbf{ OPIR(Ours)} &\textbf{42.21}	&\textbf{0.997}	&\textbf{40.54}	&0.990
         \\
         \hline
    \end{tabular}}
\end{table}

\textbf{Image Dehazing.} Table~\ref{tab:sot} reports quantitative comparisons on the SOTS~\cite{RESIDEli2018benchmarking} benchmark for image dehazing. OPIR consistently outperforms all competing methods on both the SOTS-Indoor and SOTS-Outdoor subsets. On SOTS-Indoor, OPIR achieves a PSNR of 42.21 dB, surpassing the previous best method, PGH$^2$Net~\cite{PGH2Netisu2025prior}, by 0.51 dB. On the more challenging SOTS-Outdoor dataset, OPIR attains 40.54 dB PSNR, exceeding the strongest baseline SFNet~\cite{SFNet} by 0.49 dB, while maintaining comparable SSIM performance. These consistent improvements demonstrate that OPIR more effectively removes haze while preserving fine structures and visual fidelity, highlighting its robustness and strong generalization capability across both indoor and outdoor dehazing scenarios.

\subsection{Ablation Studies}
We perform ablation studies under the all-in-one setting and conduct a step-by-step analysis by progressively incorporating the proposed modules to validate the effectiveness of each component, as summarized in Table \ref{tab:abl1}.

\begin{table*}
    \centering
    \caption{Ablation study on individual components of OPIR.}
    \label{tab:abl1}
    \begin{tabular}{ccccccc}
    \hline
        Method& Stage & Uncertainty map & Task-aware module &Multi-scale kernel &  PSNR &$\triangle$ PSNR 
         \\
         \hline
         (a) &1 & & & & 32.31 & -
         \\
         (b) &1 &&\ding{52} &  & 33.26 & + 0.95 dB
         \\
          (c) &1 & & &\ding{52}  & 32.95 & + 0.64 dB
         \\
        (d) &1 & &\ding{52} &\ding{52}  & 33.52 & + 1.21 dB
         \\
          (e) &2 & &\ding{52} &\ding{52}  & 33.77 & + 1.46 dB
          \\
           (f) &2 &\ding{52} &\ding{52} &\ding{52}  & 34.02 & + 1.71 dB
         \\
         \hline
    \end{tabular}
\end{table*}

Starting from a single-stage baseline without any proposed modules (a), the model achieves a PSNR of 32.31 dB.
By introducing the task-aware module alone (b), the PSNR increases by 0.95 dB, indicating that explicitly conditioning the restoration process on task-specific information significantly improves performance.  Incorporating the multi-scale kernel prediction module (c) yields a gain of 0.64 dB, demonstrating the effectiveness of multi-scale modeling.

When both the task-aware module and multi-scale kernel prediction are jointly applied within a single stage (d), the PSNR further improves to 33.52 dB, achieving a cumulative gain of 1.21 dB over the baseline. This result suggests that these two components are complementary, where task awareness guides the kernel prediction process to better adapt to different restoration tasks.

We further extend the model to a two-stage architecture (e), which leads to an additional improvement of 0.25 dB compared to its single-stage counterpart, highlighting the benefit of progressive refinement in image restoration. Finally, by integrating the uncertainty map into the two-stage model (f), OPIR reaches the best performance of 34.02 dB, corresponding to a total gain of 1.71 dB over the baseline. This improvement verifies that uncertainty-aware modeling effectively captures spatially varying degradation confidence, enabling more robust kernel prediction and restoration.

\begin{table}
    \centering
    \caption{The evaluation of model computational complexity. }
    \label{tab:computational}
     \resizebox{\linewidth}{!}{
    \begin{tabular}{cccc}
    \hline
         Method& Time(s) &Flops(G)  & PSNR 
         \\
         \hline\hline
         VLU-Net~\cite{VLUNetZeng_2025_CVPR} &0.743 &143  &33.17
         \\
         PromptIR~\cite{potlapalli2023promptir} &1.012 & 134 & 33.04 
         \\
         Perceive-IR~\cite{Perceive-IR10990319} &0.682 &144 &33.20 
         \\
         \hline
       \rowcolor{gray!20} \textbf{OPIR(Ours)} &\textbf{0.174} &\textbf{47} &\textbf{34.02} 
        \\
         \hline
    \end{tabular}}
\end{table}

\subsection{Resource Efficient}
We further analyze the computational efficiency of OPIR by comparing its runtime  and FLOPs with recent state-of-the-art methods. As shown in Table~\ref{tab:computational}, OPIR not only achieves leading restoration performance but also significantly reduces computational demand. In particular, it surpasses the previous best method, Perceive-IR~\cite{Perceive-IR10990319}, by 0.82 dB while requiring only 25.5\% of its inference time.

\section{Conclusion}

We propose OPIR, an efficient all-in-one image restoration framework based on physical degradation modeling. By predicting a task-aware inverse degradation operator, OPIR avoids the need for additional prompt modules or large-scale models, significantly reducing system complexity while retaining strong practical efficiency. The proposed two-stage architecture first generates an initial restoration along with an uncertainty perception map to identify challenging regions, and then refines the results under the guidance of this uncertainty information. A shared inverse operator prediction network and task-aware parameters enable effective adaptation to diverse degradation types, while convolution-based operator implementation ensures computational efficiency.
Extensive experimental results demonstrate that OPIR achieves superior performance.

\nocite{langley00}

\bibliography{example_paper}
\bibliographystyle{icml2026}

\newpage
\appendix
\onecolumn

\section{Loss Function}

Consistent with prior work, we optimize OPIR jointly in the spatial and frequency domains. The overall objective function is formulated as:
\begin{equation}
\begin{aligned}
\label{eq:loss1}
L &= \sum_{i=1}^{3}(L_{c}(\hat{J_i},\overline I)  + \delta L_{e}(\hat{J_i},\overline I) 
\\
& + \lambda L_{f}(\hat{J_i},\overline I) 
\\
L_{c} &= \sqrt{||\hat{J_i} -\overline I||^2 + \epsilon^2}
\\
L_{e} &= \sqrt{||\triangle \hat{J_i} - \triangle \overline I||^2 + \epsilon^2}
\\
L_{f} &= ||\mathcal{F}(\hat{J}_i)-\mathcal{F}(\overline I)||_1
\end{aligned}
\end{equation}
where $i$ indexes the input–output image pairs, and $\overline{I}i$ denotes the corresponding ground-truth images. $\mathcal{L}_{c}$ is the Charbonnier loss, with the constant $\epsilon$ empirically set to $0.001$ in all experiments. $\mathcal{L}_{e}$ is an edge-aware loss, where $\triangle$ denotes the Laplacian operator, encouraging accurate reconstruction of high-frequency structures. $\mathcal{L}_{f}$ represents the frequency-domain loss, in which $\mathcal{F}(\cdot)$ denotes the fast Fourier transform. The weighting factors $\lambda$ and $\delta$ are set to $0.1$ and $0.05$, respectively, following~\cite{Zamir2021MPRNet,FSNet}.

\section{More Experiments}
\label{sec:me}

\subsection{Experimental Setup}
\label{sec:mes}

\textbf{Datasets.} For the task-aligned setting, we train the deraining model using 13,712 paired clean–rain images collected from multiple datasets~\cite{Rain100,Test100,8099669,7780668}. The trained OPIR is evaluated on several benchmark test sets, including Rain100H~\cite{Rain100}, Rain100L~\cite{Rain100}, Test100~\cite{Test100}, and Test1200~\cite{DIDMDN}. For desnowing evaluation, we adopt the Snow100K~\cite{desnownet}, SRRS~\cite{JSTASRchen2020jstasr}, and CSD~\cite{HDCW-Netchen2021all} datasets. To ensure consistency with the training strategy of previous work~\cite{FSNet}, we randomly sample 2,500 image pairs for training and 2,000 images for evaluation.
For dehazing, we assess the performance of our OPIR on daytime synthetic datasets from RESIDE~\cite{RESIDEli2018benchmarking}, which include the indoor training set (ITS), the outdoor training set (OTS), and the synthetic objective testing set (SOTS). The model is trained separately on ITS and OTS, and then evaluated on the corresponding SOTS-Indoor and SOTS-Outdoor test sets, each containing 500 paired images.

For the all-in-one setting, we combine the above image restoration datasets to construct a unified training set. During evaluation, we use Test100~\cite{Test100} for deraining, Snow100K~\cite{desnownet} for desnowing, and SOTS-Outdoor~\cite{RESIDEli2018benchmarking} for dehazing.

\textbf{Training details.} All models are optimized using the Adam optimizer~\cite{2014Adam} with $\beta_1 = 0.9$ and $\beta_2 = 0.999$. The learning rate is initialized at $2 \times 10^{-4}$ and progressively reduced to $1 \times 10^{-7}$ according to a cosine annealing strategy~\cite{2016SGDR}. During training, we randomly crop $256 \times 256$ patches with a batch size of 32 and train the models for $4 \times 10^5$ iterations. Horizontal and vertical flipping are applied for data augmentation. To ensure fair comparisons, all deep learning-based baselines are fine-tuned or retrained following the hyperparameter configurations reported in their original works.

\subsection{EfDeRain+~\cite{efderainguo2025efficientderain+} vs. OPIR}
EfDeRain+~\cite{efderainguo2025efficientderain+} tackles the limitations of naive cascaded predictive filtering by introducing an uncertainty map derived from the statistics of predicted kernels, which enables a second-stage refinement that focuses on regions that are difficult to reconstruct. While this strategy is effective for single-task deraining, it remains inherently task-specific. Both predictive networks are trained exclusively for rain removal, and the uncertainty formulation is based on assumptions about kernel weight distributions that are closely coupled with rain streak characteristics. Consequently, extending EfDeRain+~\cite{efderainguo2025efficientderain+} to other restoration tasks requires retraining separate models, which restricts its general applicability.

In contrast, OPIR reformulates cascaded predictive filtering as a unified inverse operator learning framework rather than a task-dependent refinement scheme. Instead of learning independent predictive filters for different degradations, OPIR models a shared base inverse operator that is explicitly modulated by a task-aware module. Task information is injected directly at the operator level through lightweight, task-conditioned modulation, allowing a single backbone network to adapt its inverse behavior across heterogeneous restoration tasks. This design enables OPIR to capture degradation-specific characteristics while maintaining parameter efficiency and strong cross-task generalization.

A further distinction lies in uncertainty modeling. EfDeRain+~\cite{efderainguo2025efficientderain+} estimates uncertainty by averaging predicted kernel weights, motivated by the observation that rain-corrupted pixels often exhibit characteristic positive–negative weight patterns. Although effective for deraining, this formulation implicitly depends on task-specific kernel distributions. In contrast, OPIR defines uncertainty directly from the magnitude of the predicted inverse operator, reflecting the strength of local dependencies required to reconstruct each pixel. This operator-centric uncertainty formulation is task-agnostic and characterizes reconstruction difficulty from a physical perspective, making it applicable to a broader range of degradations beyond rain.

The role of the two-stage design also differs fundamentally between the two methods. In EfDeRain+~\cite{efderainguo2025efficientderain+}, cascaded filtering serves as a heuristic refinement step in which the second stage compensates for residual rain artifacts. In OPIR, the two stages correspond to progressive inverse operator estimation. The first stage produces an initial task-aware restoration together with an uncertainty map, while the second stage re-estimates the inverse operator conditioned on both the intermediate result and the uncertainty. This leads to a principled, uncertainty-guided operator refinement process rather than repeated filtering.

Finally, although both approaches employ weight-sharing multi-scale dilated filtering for computational efficiency, EfDeRain+~\cite{efderainguo2025efficientderain+} applies this strategy within a single-task deraining pipeline. OPIR integrates multi-scale filtering into a task-aware inverse operator framework, enabling adaptive receptive field selection at the pixel level while preserving low computational complexity. Moreover, OPIR generates offsets for all scales within a single loop, performs multi-scale convolution in parallel, and further reduces redundant computation through learnable fusion weights.

\begin{table}[hb]
    \centering
    \caption{Ablation study on EfDeRain+~\cite{efderainguo2025efficientderain+}.}
    \label{tab:ablefde}
    \begin{tabular}{ccc}
    \hline
         Method&  PSNR &$\triangle$ PSNR 
         \\
         \hline
         EfDeRain+~\cite{efderainguo2025efficientderain+} & 31.46 & -
         \\
         add Task-aware module& 33.11 & + 1.65 dB
        \\
         replace with Multi-scale kernel & 31.75&  + 0.29 dB
        \\
         replace with Multi-scale kernel + Task-aware module& 33.69& + 2.23 dB
         \\
         \hline
    \end{tabular}
\end{table}

To further demonstrate the advantages of our approach, we adopt EfDeRain+~\cite{efderainguo2025efficientderain+} as the baseline and progressively replace or incorporate the modules proposed in this paper. This controlled modification strategy allows us to directly assess the effectiveness of each component.

Table~\ref{tab:ablefde} reports an ablation study based on EfDeRain+~\cite{efderainguo2025efficientderain+}, analyzing the contributions of task-aware inverse operator modeling and multi-scale kernel design. Introducing the task-aware module alone leads to a substantial performance gain of +1.65~dB, which clearly surpasses the improvement obtained by replacing the single-scale kernel with a multi-scale kernel (+0.29~dB). This observation suggests that explicitly modulating the predicted inverse operator according to task characteristics plays a more crucial role than simply enlarging the receptive field.
When both components are jointly applied, the proposed method achieves an overall improvement of +2.23~dB compared to the baseline. 

\subsection{Additional Visual Results}
\begin{figure*} 
    \centerline{\includegraphics[width=1\linewidth]{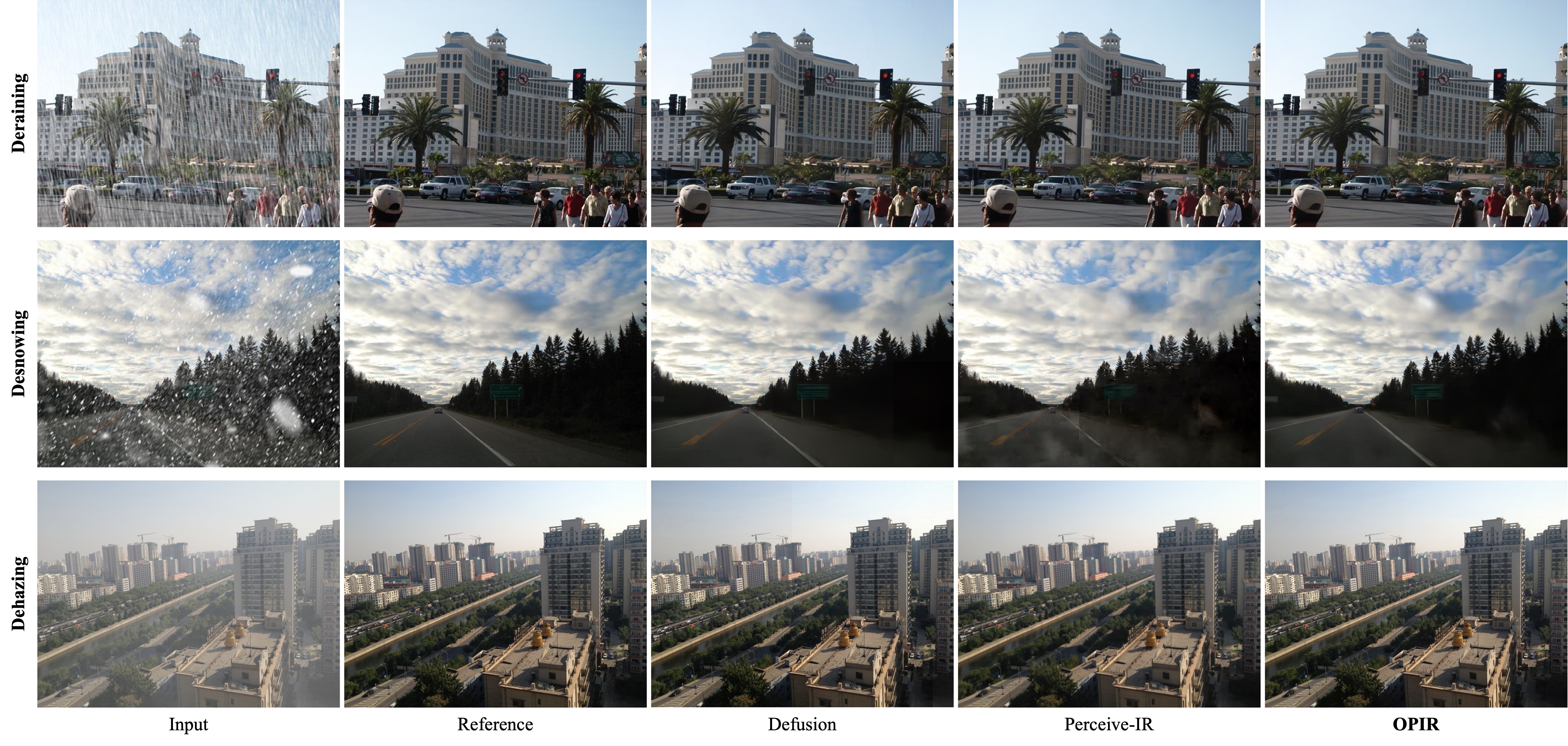}}
	\caption{Qualitative results under the all-in-one experimental setup.}
 \label{fig:suppleall-in-one}
\end{figure*}

\begin{figure*} 
    \centerline{\includegraphics[width=1\linewidth]{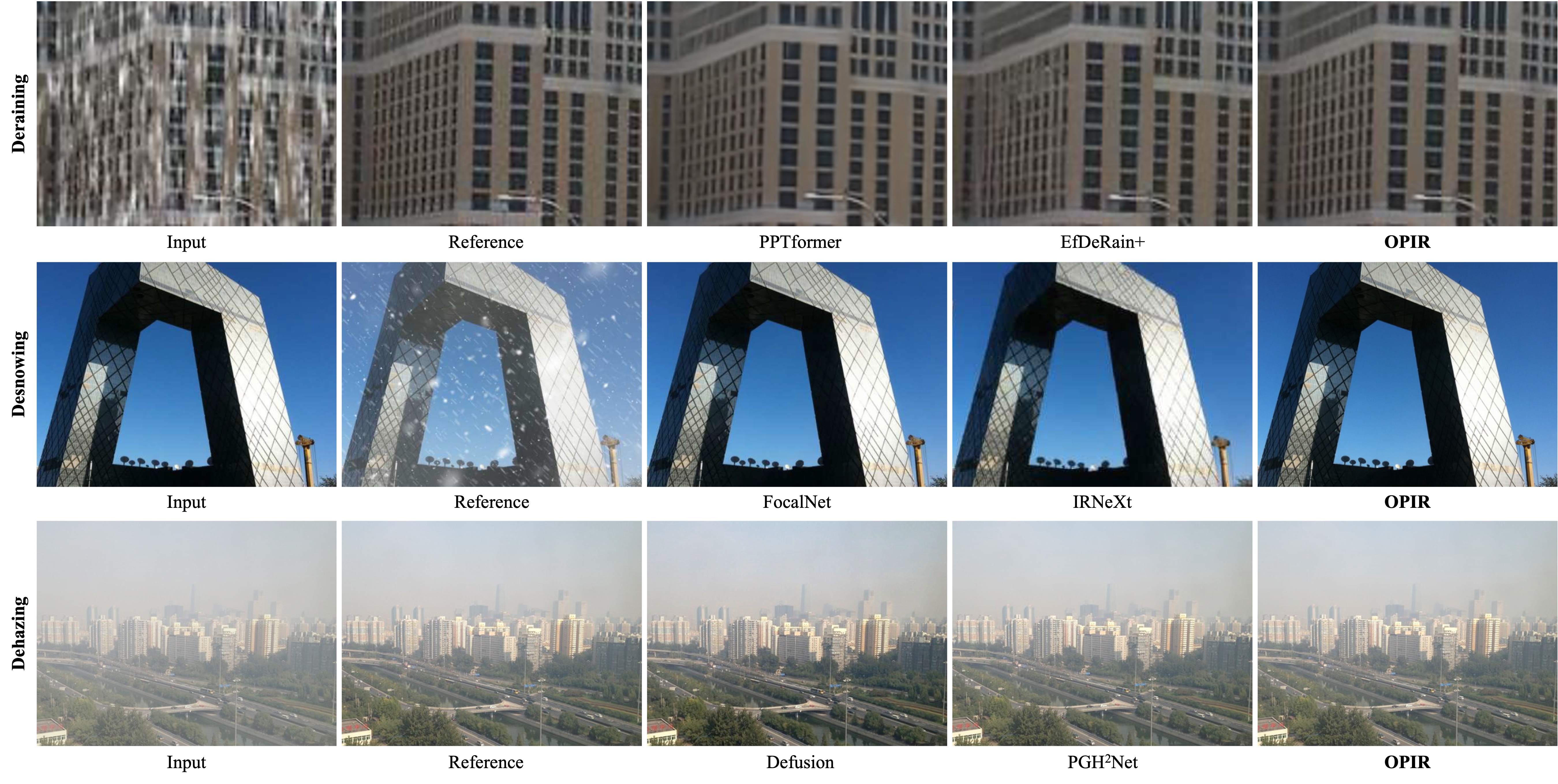}}
	\caption{Qualitative results under the task-aligned experimental setup.}
 \label{fig:supplesig-in-one}
\end{figure*}

Figure~\ref{fig:suppleall-in-one} presents qualitative comparisons of our method under the all-in-one experimental setting, where a single unified model is used to handle multiple degradation types. Figure~\ref{fig:supplesig-in-one} shows the corresponding visual results under the task-aligned setting, in which the model is explicitly adapted to each specific degradation. As can be seen, our approach effectively removes diverse degradations across both settings, while consistently reconstructing images with sharp structural details, natural textures, and visually pleasing overall appearance. These results demonstrate the robustness and strong generalization capability of the proposed method under different restoration paradigms.

\end{document}